# Multilingual, Multimodal Pipeline for Creating Authentic and Structured Fact-Checked Claim Dataset

**Z. Melce Hüsünbeyi**[1,2], **Virginie Mouilleron**[2], **Leonie Uhling**[1], **Daniel Foppe**[1], **Tatjana Scheffler**[1], **Djamé Seddah**[2]
[1]Ruhr-Universität Bochum, [2]Inria Paris
melce.husunbeyi@rub.de

**Abstract**

The rapid proliferation of misinformation across online platforms underscores the urgent need for robust, up-to-date, explainable, and multilingual fact-checking resources. However, existing datasets are limited in scope, often lacking multimodal evidence, structured annotations, and detailed links between claims, evidence, and verdicts. This paper introduces a comprehensive data collection and processing pipeline that constructs multimodal fact-checking datasets in French and German languages by aggregating ClaimReview feeds, scraping full debunking articles, normalizing heterogeneous claim verdicts, and enriching them with structured metadata and aligned visual content. We used state-of-the-art large language models (LLMs) and multimodal LLMs for (i) evidence extraction under predefined evidence categories and (ii) justification generation that links evidence to verdicts. Evaluation with G-Eval and human assessment demonstrates that our pipeline enables fine-grained comparison of fact-checking practices across different organizations or media markets, facilitates the development of more interpretable and evidence-grounded fact-checking models, and lays the groundwork for future research on multilingual, multimodal misinformation verification.

## 1. Introduction

The global proliferation of misinformation across languages, platforms, and modalities has intensified the need for robust, real-world, and continuously updated resources for automated fake-news detection. Yet most available benchmarks remain English-dominant, text-centric, and static, offering sparse annotations and only limited coverage of the images and videos that increasingly drive the propagation of falsehoods (Akhtar et al., 2023; Abdali et al., 2024). Meanwhile, fact-checking organizations counter misinformation through systematic, journalism-based workflows, collecting and cross-verifying evidence from diverse sources, contextualizing claims, and articulating evidence-based justifications (Cazzamatta, 2025a; Johnson, 2024). Despite advances in automated fact-checking, current research remains constrained by the lack of datasets that reflect these practices with sufficient depth and diversity.

Most existing systems apply stance detection to documents retrieved via relevance scoring, provide no local, checkable pointers (e.g., text spans, timestamps), and over-rely on secondary sources (e.g., Wikipedia, news articles) instead of probative primary materials such as official records or expert testimony (Warren et al., 2025; Schlichtkrull et al., 2023). A significant reason for this limitation is that automated systems treat all retrieved documents as interchangeable text, whereas fact-checkers rely on functionally distinct sources each providing a different reason to accept or reject a claim. Automated systems cannot replicate how fact-checkers prioritize primary evidence without encoding its context-dependency, such as the distinct need for court documents, official statements, or multimedia (Warren et al., 2025).

To address these gaps, we propose a multistage, scalable pipeline for constructing multilingual, multimodal fact-checking datasets, which centered on French and German claims that incorporate real-world data from fact checking organizations. The pipeline aggregates ClaimReview formatted claim statements and corresponding debunking articles from publisher sites, systematically links text, images, and videos to each verified claim, and normalizes heterogeneous rating schemes. This enables us to move beyond static corpora and support rigorous real-world evaluation of both text-only and multimodal large language models (LLMs) for various fact-checking related tasks.

In contrast to the coarse evidence handling typical of existing automated fact-checking systems, our pipeline centers on a fine-grained evidence schema that classifies evidence according to the role it plays in verification. This design is motivated by journalism and fact-checking research that treats verification as a process of source criticism and triangulation rather than simple document matching (Steensen et al., 2022; Godler and Reich, 2017). It also draws on empirical work showing that professional fact-checkers rely on functionally distinct evidence types that vary systematically by organizational context and verification complexity (Cazzamatta, 2025b). Accordingly, we refine coarse-grained evidence typologies into more specific epistemic categories (e.g., expert testimony; official records; media records), while preserving local, checkable pointers to evidence items (e.g., references, timestamps) where available, thereby enabling cross-source validation. Using zero-shot LLMs, we identify, extract and categorize evidence items from published debunking articles under this schema, and then generate justifications that explicitly reason over the categorized evidence and align with each claim's verdict. We evaluate the system outputs for evidence extraction and justification generation in textual and multimodal settings with G-Eval (Liu et al., 2023), along three criteria: correctness, coherence, and completeness. The result is a reliable dataset that supplies fine-grained, category-based annotations aligned with fact-checking rationales across text, image, and video, and pairs each claim with a justification grounded in those categories. Since evidence types represent key dimensions of objectivity and credibility in fact-checking practice (Cazzamatta, 2025b), by consistently encoding them across cases, our proposed pipeline enables cross-cultural and cross-organizational analyses of fact-checking practices, revealing how verification strategies vary across media systems and contexts.

Our main contributions are as follows:

- A fully scriptable, reproducible pipeline for constructing multilingual, multimodal fact-checking datasets enriched with structured metadata and evidence annotations.
- Two new, up-to-date large-scale datasets for French and German containing claims, debunking articles, visual media, and evidence categorized according to journalistic verification practices.
- An analysis of evidence category usage across organizations to highlight institutional differences in verification practices across Europe.
- A comprehensive benchmarking of LLM and multimodal LLM based systems for evidence extraction and justification generation along with LLM-as-a-judge approach.

## 2. Related Work

Despite the global proliferation of misinformation, the availability of robust multilingual and multimodal fact-checking datasets remains severely restricted, particularly for non-English languages (Nielsen and McConville, 2022; Panchendrarajan and Zubiaga, 2024). While large-scale multilingual corpora such as MultiClaim (Pikuliak et al., 2023), X-Fact (Gupta and Srikumar, 2021), and NewsPolyML (Mohtaj et al., 2024) encompass French and German, they offer a limited volume of claims and lack diverse local sources. Furthermore, these resources are text-centric, neglecting the visual dimensions of misinformation.

Existing French-language fact-checking resources typically address narrow segments of the verification landscape. StatCheck (Balalau et al., 2022) focuses exclusively on statistical claims verified against official datasets (e.g., INSEE, Eurostat). LIAR-New (French) (Pelrine et al., 2023) relies on translated and GPT-4-augmented statements from the American PolitiFact dataset; consequently, it is a text-only resource that fails to reflect the authentic French media ecosystem or provide native evidence trails. OBSINFOX (Icard et al., 2024) provides 100 French press documents from 17 sources identified as unreliable by NewsGuard and Conspiracy Watch, annotated across eleven labels capturing linguistic and content-level features of misleading content; its focus remains on the linguistic profile of unreliable sources, without claim-level verification. Bachelot et al. (Bachelot et al., 2024) construct a dataset of over one million French tweets across fifteen events with propagation graphs and cascade trees; veracity labels are assigned at the event level by French fact-checking organizations, with individual tweets inheriting the event-level label. Although FactCheckBureau (Balalau et al., 2024) provides a large-scale corpus of claim reviews with French coverage limited to 1.2K aligned tweet–article pairs, as a retrieval-oriented system, verdict labels remain heterogeneous across sources.

German-language datasets face significant limitations regarding domain specificity and claim-level verification. FANG-COVID (Mattern et al., 2021) provides pandemic-related news and propagation data, yet articles are labeled based on the reliability of their source rather than the veracity of individual claims. DeFaktS (Ashraf et al., 2024) supplies over 100,000 Twitter posts with fine-grained, taxonomy-based labels across semantic, psychological, authenticity, and content-type dimensions, grounding its schema in linguistic framing rather than external evidence verification. While HealthFC (Vladika et al., 2024) pairs 750 health-related claims with medical literature evidence, it remains text-centric and confined to the health domain. GERMA (Carrella and Miani, 2025) aggregates over 237,000 articles from unreliable sources, yet captures source-level unreliability rather than verified false claims.

Recent surveys emphasize that most available resources fail to capture the complexity of multimodal misinformation, wherein images and videos play a central role in driving the spread and perceived credibility of false claims (Nielsen and McConville, 2022; Geng et al., 2024). While datasets like MMFakeBench (Liu et al., 2024) and MMOOC (Xu et al., 2024) have advanced image text fact-checking, they rely heavily on synthetic construction, generating samples by swapping or manipulating visual and textual modalities. English-language research has consequently shifted toward real-world multimodal benchmarks, such as AVERIMATEC (Cao et al., 2025), which explicitly encodes reasoning paths using question-answer pairs drawn from authentic fact-checking articles. Although Rayar (Rayar, 2024) aligns French political statements with broadcast video using speech and face recognition, its scope is restricted to political discourse.

Despite the fact that many claims require cross-source evidence retrieval and evaluation, automated fact-checking systems remain limited in capturing the complexity of real-world verification practice. Drawing on interviews with professional fact-checkers, Warren et al. (2025) find that practitioners consistently prioritize primary sources, such as official records, datasets, and firsthand statements, while expressing broad skepticism toward automated tools on the grounds that such tools neither replicate this source assessment process nor adequately explain their outputs. Furthermore, Cazzamatta's (2025b) comparative research across European fact-checking organizations has highlighted that evidence in fact-checking is not monolithic; rather, it encompasses a range of types. This study shows that the types of evidence employed vary significantly by media system and organizational context, with more robust, "all-inclusive" approaches combining documentation, digital forensic tools, and statements found in contexts of higher journalistic professionalism and linked to higher verification complexity.

By collecting real-world claims for French and German and systematically organizing the evidence sources used by fact-checking organizations into fine-grained categories, complemented by aligned visual data and structured metadata, our approach addresses the gap between automated fact-checking and journalistic verification practice.

## 3. Data collection pipeline

We designed a multi-stage pipeline (see Fig. 1) for constructing a multilingual, multimodal fact-checking dataset by aggregating recent fact-checking articles and enriching them with structured metadata and aligned media. In step ①, our primary sources were ClaimReview feeds, obtained through the fact check Google claim review API[1] and the original publisher websites hosting the full debunking articles. The ClaimReview format is a standardized schema widely adopted by fact-checking organizations to mark up verified claims for search engines. By querying the API per publisher and language, we collected claim statements incorporating canonical keys such as publisher, date, claim, review URL, review title, language, and the original rating provided by the fact-checker.

[1] https://developers.google.com/search/docs/appearance/structured-data/factcheck

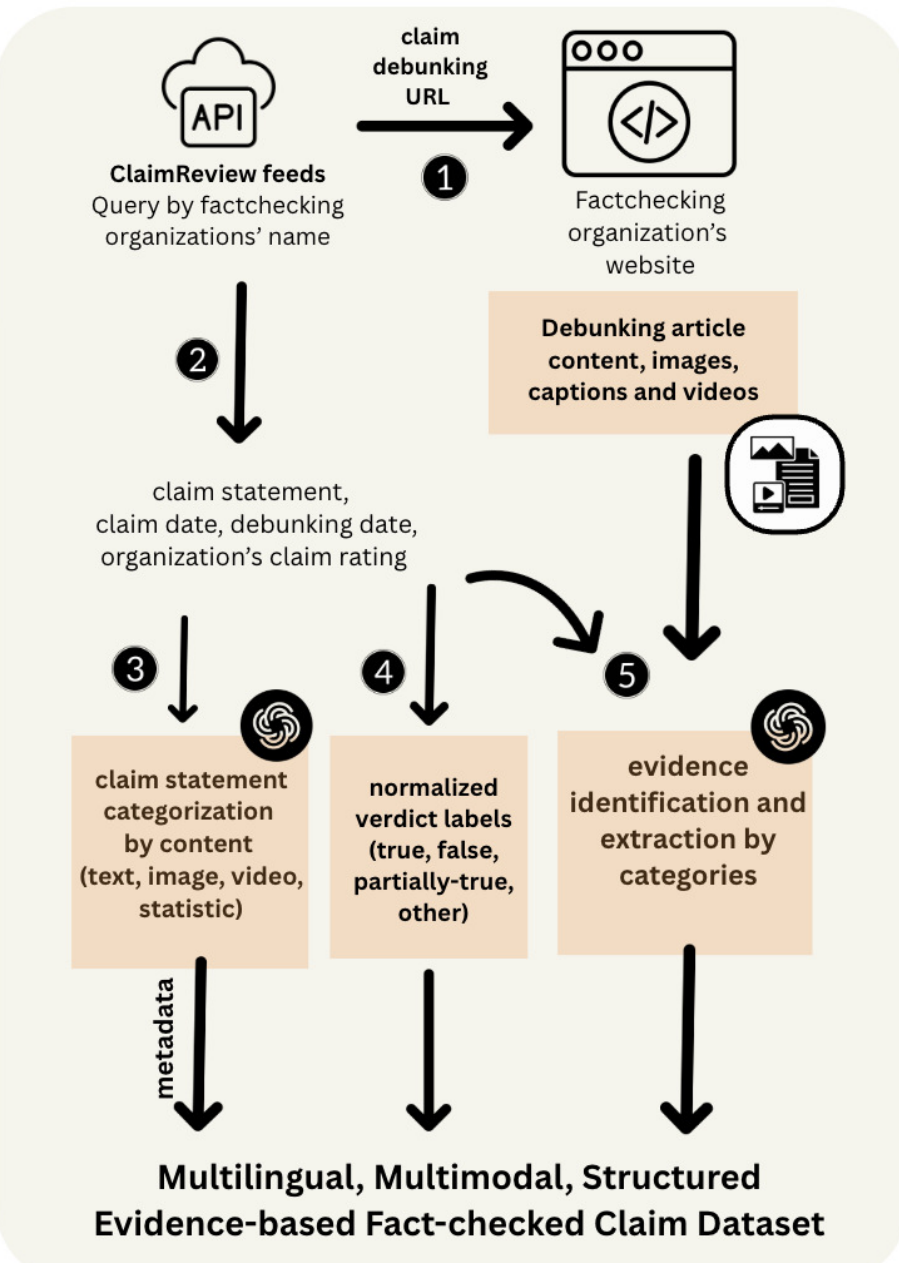

Figure 1: Pipeline overview.

In step ②, for incorporating contextual information and evidence statements of reviewed claims, we scraped the debunking articles based on the review URL information from the corresponding claim statement. Because fact-checking organizations differ in how they present and structure their content, this step required organization-specific scraping strategies. Dedicated scraping adapters were developed for each organization's website to reliably extract the article content along with any embedded images, their captions and videos.

We then performed metadata enrichment to standardize and augment the collected information. In step ③, we used a few-shot–prompted GPT-4o model (OpenAI et al., 2024) to categorize claim sources based on concepts appearing within the claim sentences, image, video, and statistical data, which further improves the dataset's utility for multimodal fact-checking research. In step ④, given the heterogeneity of rating schemes across fact-checking organizations, ranging from binary labels (True/False) to numeric scales or free-text author conclusions, standardizing truthfulness ratings was essential to ensure consistency in our dataset. To achieve this, we mapped the varying ratings deterministically into verdict labels that have been defined by ClaimsKG (Tchechmedjiev et al., 2019) (i.e., True, False, Mixture[2], Other). We inherited and extended their predefined lookup table that aligns each organization's labeling convention with a unified label set, developing language-specific normalization strategies based on lookup-based matching, and regular expressions targeting key indicator phrases and deterministic scoring functions where applicable.

[2] Throughout this paper, we use *partially-true* to refer to this category.

Finally, step ⑤ concerns evidence extraction. To ensure that our dataset reflects the principles and workflows of professional fact-checking, we adopt a journalistic approach to evidence representation. In journalistic practice, verification is not limited to matching claims with isolated facts but involves a systematic process of gathering, cross-checking, contextualizing, and synthesizing information from a diverse range of sources. This multi-source strategy supports transparency, enables triangulation, and situates claims within their broader social and political contexts. To capture this evidentiary logic, we categorize the available evidence using a fine-grained schema grounded in journalistic verification as represented in debunking articles. Our evidence schema draws on distinctions between official documents, data and statistics, digital sources, oral records, and expert authority (Baker and Fairbank, 2022), and also builds on Cazzamatta (2025b)'s broader distinction between documentation, statements, and digital forensic tools in European fact-checking practice. We refine these broader evidence sources into six operational categories defined by the epistemic role an evidence item serves in the debunking article: Expert Testimony, Quantitative Data and Statistics, Official Records and Legal/Policy Documents, Media Record (Mass or Social), Multimedia Evidence, and Eyewitness and Firsthand Accounts. This role-based design allows us to distinguish evidence types that may share a similar textual or material form but contribute differently to verification. For example, expert testimony is distinguished from eyewitness and firsthand accounts according to whether credibility derives primarily from recognized expertise or from direct personal experience. We also treat multimedia objects as evidence when the image, video, audio clip, screenshot, or chart is itself cited as proof, while digital forensic tools are treated as verification methods (Bjerknes, 2022). In this way, our schema provides an operational representation of the evidentiary roles made explicit in published debunking articles and supports LLM-based extraction and dataset construction.

**Expert Testimony** refers to statements or analyses whose credibility comes from the recognized knowledge, training, or authority of a professional in a relevant field. This includes interviews, commentary, and judgments made by professors, doctors, scientists, engineers, lawyers, and other licensed specialists, as well as official spokespersons acting explicitly in an expert capacity.

**Quantitative Data and Statistics** covers evidence in the form of measurable figures used as proof. This includes percentages, rates, margins of error, poll results, survey data, financial statements, budgets, and model estimates such as forecasts or projections.

**Official Records and Legal/Policy Documents** are formal and authoritative materials issued by governments, institutions, or organizations in an official capacity. This category encompasses laws, regulations, treaties, contracts, court filings, judicial rulings, governmental and institutional reports, etc.

**Media Record (Mass or Social)** covers publications from journalism and non-governmental social media accounts that serve as evidence of the publication or circulation of a claim or piece of information through mass or social media, such as news articles, editorials, blogs, newsletters, online forums, and social media posts.

**Multimedia Evidence** includes all forms of visual, audio, or interactive files used as a proof. This covers photos, videos, audio recordings, screenshots of posts, charts, or documents, as well as embedded social media posts when the media object itself is the cited evidence. Maps, diagrams, and graphs presented as images also qualify.

**Eyewitness and Firsthand Accounts** captures evidence whose credibility derives from direct personal observation or lived experience of the event in question, as distinct from institutionally recognized expertise.

Structuring evidence in this way has two main benefits. First, it models the epistemic diversity of real-world fact-checking, enabling downstream systems to learn from authentic verification practices rather than simplistic text-matching. Second, it enhances interpretability and transparency, allowing researchers and models alike to trace the reasoning from claim through evidence to verdict. Ultimately, this categorization captures the investigative reasoning central to journalistic fact-checking and lays the foundation for building more accurate and trustworthy fact-checking systems.

Overall, our pipeline supports multilingual and multimodal fact-checking by integrating category-based evidence texts, images, and videos into a unified dataset. Moreover, because each stage, from API harvesting and site-level scraping to label normalization and media alignment, is fully scriptable and executable end-to-end, this approach facilitates the creation of up-to-date real-world datasets. To support reproducibility, we publicly release the dataset, the full set of LLM prompting templates, and all pipeline scripts at `https://github.com/zmelce/factcheck.git`.

## 4. Novel multilingual, multimodal factchecking datasets

We applied this pipeline to media markets in two languages, French and German, producing two large-scale, multilingual, and multimodal fact-checking datasets. Each dataset integrates structured claim reviews, enriched metadata, and aligned visual evidence (images, videos, and their captions). The resulting resources provide detailed annotations not only of the claim statements and their verdicts but also of the underlying evidence, categorized according to journalistic verification practices.

| publisher | FALSE | TRUE | partially-true | other |
|---|---|---|---|---|
| factuel.afp.com | %39,75 | %0,17 | %1,16 | %0,19 |
| tf1info.fr | %13,17 | %5,05 | %2,03 | |
| 20minutes.fr | %9,81 | %1,30 | %0,31 | |
| lemonde.fr | %8,53 | %0,10 | %0,04 | |
| francetvinfo.fr | %3,66 | %2,32 | | |
| dpa-factchecking.com | %5,67 | | %0,08 | %0,02 |
| defacto-observatoire.fr | %2,09 | %0,14 | %0,14 | %0,04 |
| fasocheck.org | %1,37 | %0,08 | | |
| liberation.fr | %1,28 | | | |
| numerama.com | %0,85 | %0,04 | %0,10 | |
| science.feedback.org | %0,27 | %0,02 | | |
| observalgerie.com | %0,25 | | | |

Figure 2: French dataset distribution by fact checking organizations and verdict labels.

| publisher | FALSE | partially-true | TRUE | other |
|---|---|---|---|---|
| correctiv.org | %22,08 | %14,71 | %1,72 | %0,82 |
| dpa-factchecking.com | %24,73 | %2,08 | %0,06 | %0,56 |
| faktencheck.afp.com | %13,64 | %0,08 | | %0,03 |
| presseportal.de | %5,91 | %0,56 | %0,11 | %0,2 |
| br.de | %3,01 | %0,68 | %0,28 | |
| apa.at | %2,28 | %0,93 | %0,31 | |
| tagesschau.de | %1,63 | %0,22 | %0,03 | |
| stern.de | %1,29 | | %0,03 | |
| volksverpetzer.de | %0,84 | %0,03 | | |
| rtl.de | %0,45 | %0,03 | %0,03 | %0,03 |
| www1.wdr.de | %0,06 | %0,25 | %0,31 | |

Figure 3: German dataset distribution by fact checking organizations and verdict labels.

Our French corpus covers 5,170 claims published in the ClaimReview format between 2014 and 2025 by dedicated fact-checking organizations and newspaper fact-checking desks (Fig. 2). Each claim is paired with structured metadata, the full debunking article scraped from the source site, and any associated visual evidence when available. Similarly, our German corpus covers 2017–2023 and contains 3,555 claims, each linked to its corresponding debunking article, metadata, and available visual materials (Fig. 3). As shown in the figures, we incorporated claim reviews from multiple sources over a broad time span and across diverse topics to make the dataset more representative and inclusive.

| Claim content | Text | Image | Video | Statistic |
|---|---|---|---|---|
| French claimReview data | 74.6% | 9% | 8.4% | 8% |
| German claimReview data | 75.2% | 11.5% | 6.4% | 6.9% |

Table 1: Claim statements by their content type.

The verdict labels across both datasets were deterministically normalized to ensure consistency, since fact-checking organizations employ widely varying rating schemes. For the French corpus, we applied a rule-based approach: a lookup table maps standardized labels used consistently across French organizations to their corresponding verdict categories, including minor surface variations of these labels, and a set of regex patterns targeting sentence-initial indicators maps remaining explicit phrases to their verdict class. For the German corpus, the availability of the human-annotated ClaimReview-DE dataset introduced in our prior work (Hüsünbeyi et al., 2025) enabled a more systematic strategy. That dataset provides manually verified verdict mappings for heterogeneous ratings from several German fact-checking organizations, serving as annotator-validated ground truth. Drawing on this resource, we derived a layered rule-based system: prioritized regex patterns targeting explicit verdict vocabulary and sentence-initial verdict phrases; complemented by deterministic scoring functions that aggregate weighted linguistic signals.

Together, these two datasets enable systematic cross-organizational comparisons of fact-checking practices, supporting research on epistemic strategies, diversity of verification approaches across media ecosystems, and benchmarking of automated fact-checking systems under realistic, language- and context-specific conditions (see Table 1, details in Figs. 4, 5, and Section 5).

## 5. Evidence extraction and justification generation

Automated fact-checking systems must accomplish two key tasks: first, extracting the relevant evidence from a heterogeneous information space, and second, generating a structured justification that links this evidence to the final verdict. Our work addresses this dual challenge by leveraging our novel dataset, which for the first time provides detailed, category-based annotations of evidence and their alignment with fact-checking rationales across text, image, and video modalities. This rich annotation schema enables us to model the evidentiary reasoning processes that underlie professional fact-checking, making it possible to train and evaluate systems not only on claim verification outcomes but also on the quality and structure of their justifications. By replicating this critical aspect of journalistic verification practice (see example in Fig. 6), our dataset lays the groundwork for building fact-checking models that reason more transparently, explain their decisions, and align more closely with human fact-checkers' methods.

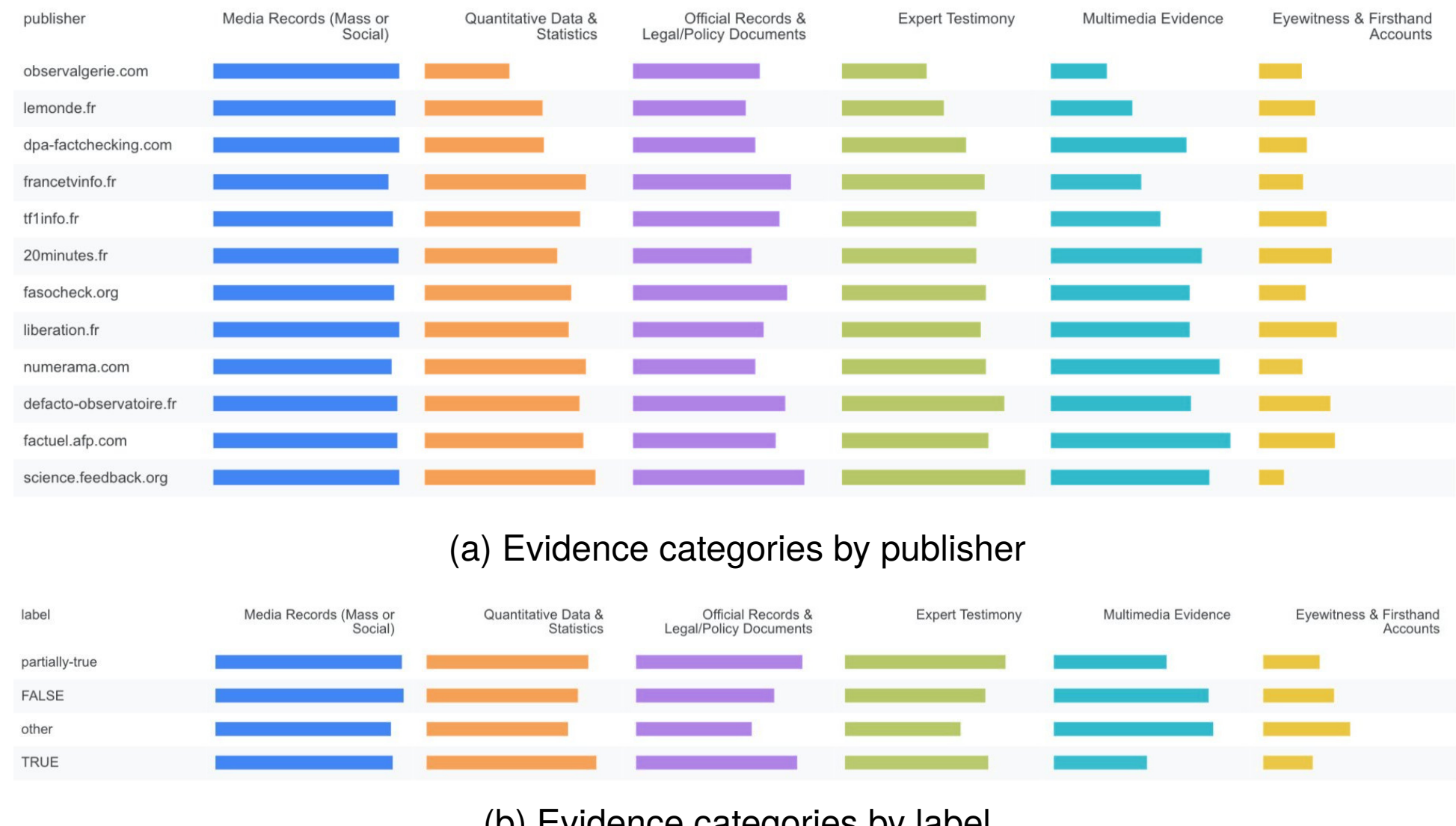

(a) Evidence categories by publisher

(b) Evidence categories by label

Figure 4: French dataset – evidence category distribution

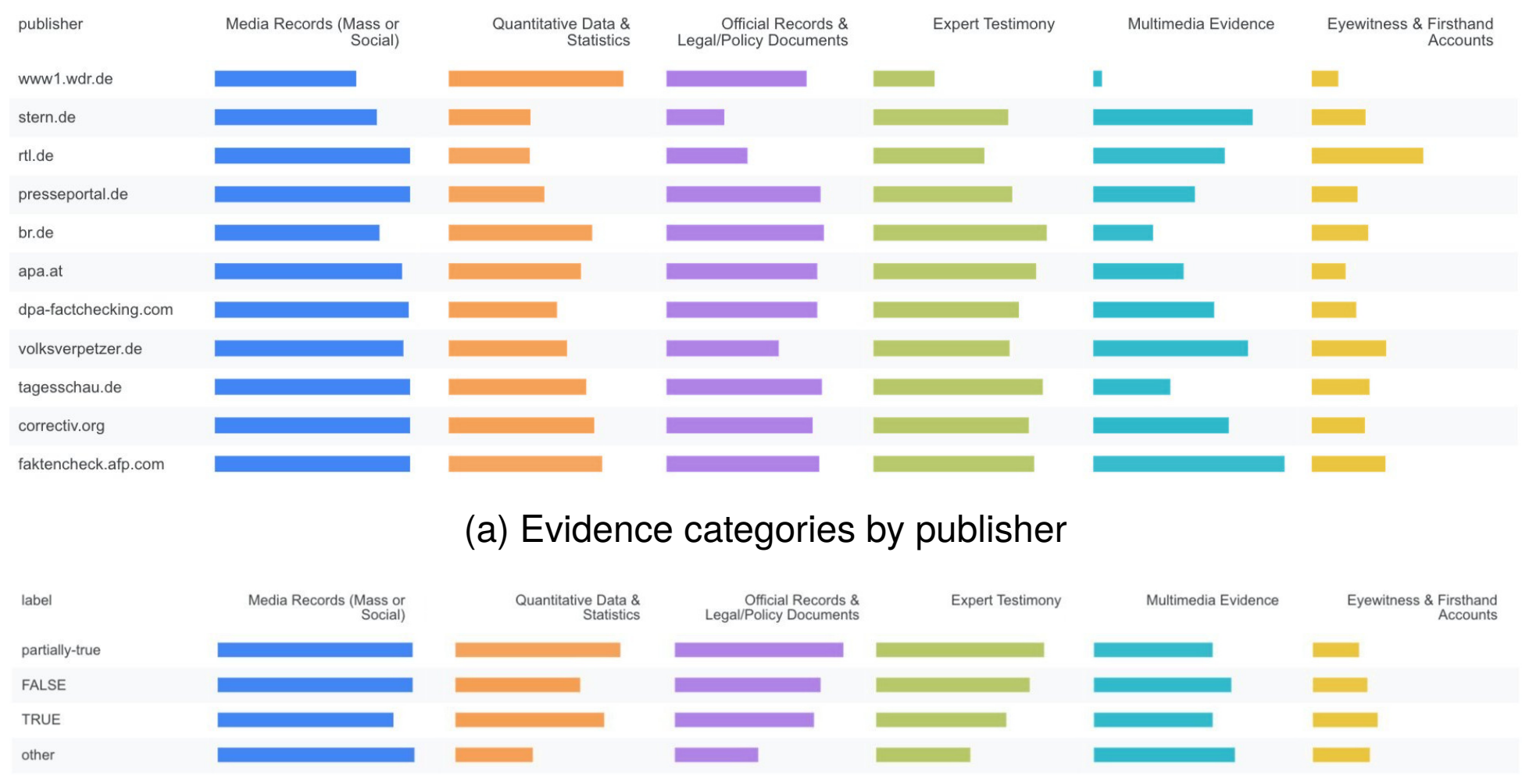

(a) Evidence categories by publisher

(b) Evidence categories by label

Figure 5: German dataset – evidence category distribution

### 5.1. Evidence extraction

We use Gemini-2.5-Pro (Comanici et al., 2025), Llama3.3-70B (Grattafiori et al., 2024) and Qwen2.5-72B (Qwen et al., 2025) models for evidence extraction along with prompting. Each call is made with a low-temperature configuration (temperature = 0.2) to favor determinism and reduce paraphrastic drift. Our proposed system ingests French and German claims alongside their corresponding debunking articles and, for each instance, composes a single, structured prompt that (i) restates the claim, (ii) provides the full article text, and (iii) embeds a rigorously specified task block defining the six evidence categories and associated output constraints. The prompt enforces article-only grounding (excluding external knowledge), fidelity to source via quotation or close paraphrase, and the use of stable inline source locators for traceability, where available.

We then examine the data distribution of the LLM-extracted evidence types across publishers and veracity labels. Figures 4 and 5 visualize the relative evidence profiles within each language dataset, while full statistical results including $\chi^2$ test results and standardized residuals (SR) are reported in Appendix Tables 5–7. In the French claim dataset (Fig. 4a), *science.feedback.org* shows the highest rates of expert testimony, official records, and quantitative evidence among all French publishers, while contributing the lowest eyewitness material. By contrast, *factuel.afp.com* shows the highest multimedia usage. News outlets such as *francetvinfo.fr* and *tf1info.fr* favor quantitative and official records, with considerably lower multimedia usage. Regarding verdict labels (Fig. 4b), *partially-true* claims exhibit the strongest reliance on documentary evidence, with expert testimony and official records reaching the highest proportions among all categories. *TRUE* items show the highest quantitative evidence alongside the lowest multimedia usage. The *other* category yields the lowest proportions of expert and official records while showing the highest eyewitness and multimedia values.

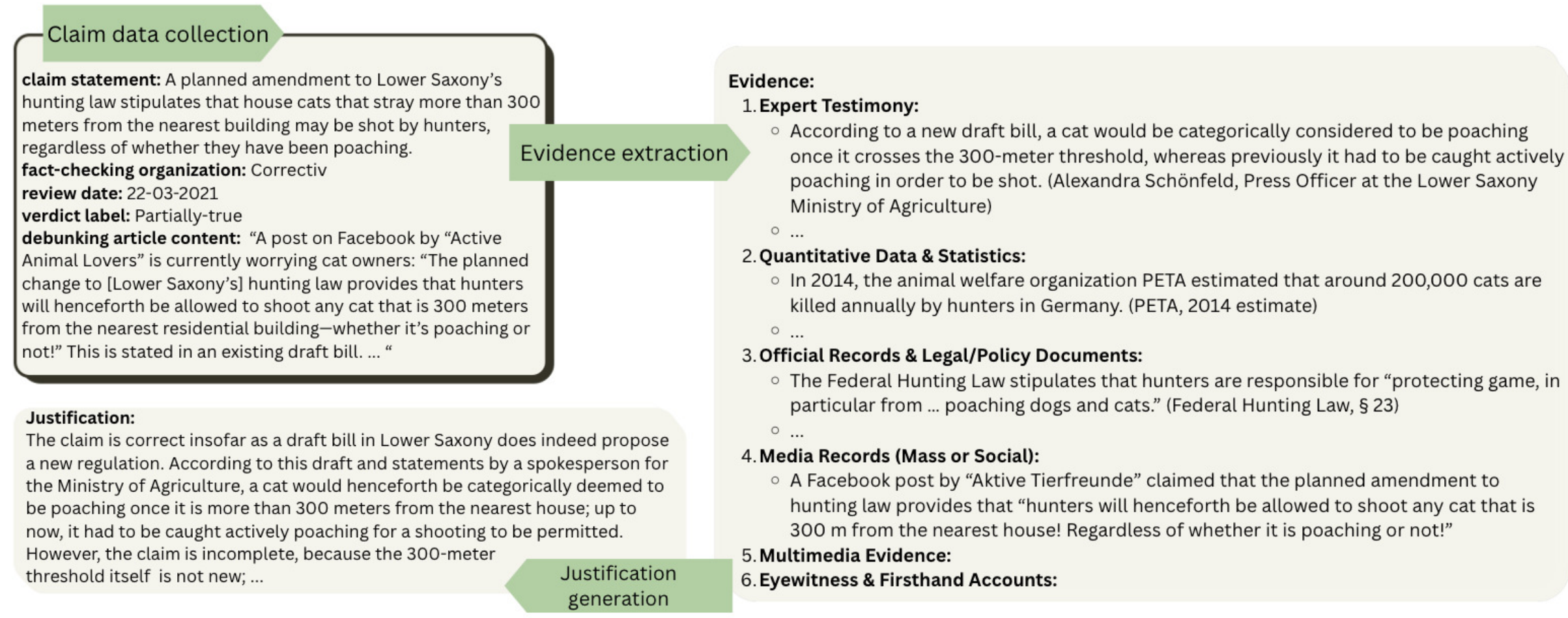

Figure 6: Evidence extraction and justification generation example (translated from German).

In the German claim dataset (Fig. 5a), *br.de*, *tagesschau.de*, and *apa.at* display high expert testimony alongside significantly low multimedia usage. Regarding verdict labels (Fig. 5b), *partially-true* claims again show the highest proportions of expert, official, and quantitative evidence, with significantly lower multimedia and eyewitness usage. The *other* category reaches the lowest proportions across these documentary evidence types, while *FALSE* labeled claims maintain high media records and multimedia proportions.

Comparing the two language datasets, French publishers demonstrate significantly greater use of quantitative evidence (SR = +4.46), multimedia evidence (SR = +3.41), and eyewitness accounts (SR = +4.71). German publishers show higher expert testimony usage (79.3% vs. 75.1%, SR = +1.69), though this difference is modest in residual analysis (see Appendix Table 5). In aggregate, these patterns indicate that (i) publisher specialization can shape evidence usage, with science-focused outlets privileging expert and quantitative sources; (ii) across languages, *partially-true* verdicts draw most heavily on documentary evidence (expert/official/quantitative), while the *other* category exhibits the lowest representation in these documentary evidence types.

## 5.2. Justification generation

We generated textual justifications in two systems using both LLMs: Llama3.3-70B and Qwen2.5-72B and multimodal LLMs: Gemini-2.5-Pro and Qwen2.5-VL-72B (Bai et al., 2025) for French and German claim statements in a zero-shot, low-temperature (temp. = 0.2) setting. Our proposed prompting systems ingest the claim, its mapped verdict label, and the previously extracted evidence. This relies solely on the extracted evidence (no external facts or speculation) with unambiguous and consistent references to specific evidence items, articulates the key reasoning steps, including addressing salient counter-evidence where present, and avoids gaps or unsupported assertions.

The first system is based on only textual input along with a structured prompt that (i) restates the claim, (ii) supplies the category based evidence text, and (iii) appends a tightly specified task block defining a single required output section ("Justification:") enforcing clear reasoning to trace a transparent progression from the claim through the cited evidence to a label-consistent conclusion.

We developed a second system that includes visual (images, associated captions, video frames and timestamps) alongside textual evidence. To ensure explicit use of visual evidence, the system instruction includes a "multimodal mode" that (i) requires taking into account image captions and video frames into the reasoning, (ii) demands grounded references to identifiable visual cues, and (iii) enforces timestamp information for video frames. We extracted compact, semantically rich video summaries from claim-relevant videos to serve as computationally efficient inputs for multimodal large language models. Our frame extraction pipeline initially samples candidate frames at regular intervals and embeds them for semantic comparison. By adapting the clustering and redundancy elimination methods from Tan et al. (2024), we systematically discard uninformative frames, yielding a highly diverse set of keyframes and corresponding timestamps.

| LLM-as-a-judge: G-Eval with GPT-4o | | French claimReview data | | | German claimReview data | | |
|---|---|---|---|---|---|---|---|
| | | coherence | correctness | completeness | coherence | correctness | completeness |
| Gemini2.5-pro | Evidence | **80.79** | **82.20** | **81.60** | **80.35** | **82.61** | **82.92** |
| | Justification | **87.63** | **94.67** | **89.46** | **87.55** | **93.15** | **89.67** |
| Qwen2.5-72B | Evidence | 75.60 | 71.44 | 69.24 | 75.90 | 76.46 | 73.68 |
| | Justification | 87.30 | 93.24 | 88.30 | 87.00 | 92.31 | 89.10 |
| LLama3.3-70B | Evidence | 78.73 | 71.28 | 71.03 | 78.01 | 72.22 | 72.52 |
| | Justification | 86.42 | 91.72 | 87.31 | 85.68 | 89.80 | 87.10 |

Table 2: Comparison of G-Eval scores for LLM results of evidence extraction and generated justification based on evaluation criteria: Coherence, Correctness, Completeness.

| LLM-as-a-judge: multimodal G-Eval with GPT-4o | | French claimReview data | | | German claimReview data | | |
|---|---|---|---|---|---|---|---|
| | | coherence | correctness | completeness | coherence | correctness | completeness |
| Gemini2.5-pro | text - only | 87.31 | 96.64 | 89.26 | 88.25 | 95.38 | 90.02 |
| Qwen2.5-VL-72B | text - only | 87.70 | 96.39 | 90.18 | 87.23 | 96.42 | 90.69 |
| Gemini2.5-pro | text + image + video frames | 88.15 | 97.07 | 89.86 | 88.21 | 96.97 | 90.76 |
| Qwen2.5-VL-72B | text + image + video frames | 86.73 | 95.34 | 88.10 | 85.76 | 94.40 | 88.03 |

Table 3: Comparison of multimodal G-Eval scores for multimodal LLM results of generated justification based on evaluation criteria: Coherence, Correctness, Completeness.

## 6. Evaluation and Results

### 6.1. Evaluation criteria

Comprehensive evaluation and comparison are essential for constructing high-quality LLM benchmarks that ensure reliability and effectiveness. Accordingly, we reviewed existing evaluation frameworks and adopted three focused criteria: *correctness* (Irnawan et al., 2025), *coherence* (Kim et al., 2024; Braun et al., 2024), and *completeness* (Braun et al., 2024). *Correctness* quantifies the factual accuracy of both the claim verdict and its supporting evidence; *coherence* captures the internal logical consistency of the justification; and *completeness* assesses whether all required evidence categories are covered and the reasoning is fully articulated. This evaluation phase yields both qualitative judgments and quantitative scores on LLM outputs.

### 6.2. LLM-as-a-judge: G-Eval

We adopted the G-Eval LLM-as-a-judge framework (Liu et al., 2023), which has been shown to correlate more closely with human judgments than traditional reference-based metrics for evaluation of LLM outputs. In our experiments, we instructed GPT-4o to reason step-by-step over the three mentioned criteria. Unlike coarse-grained rubrics, we use fine-grained, task-specific rubrics with explicit scoring and penalty items that the judge applies via predefined evaluation steps. This design evaluates both extracted evidence and generated justifications, yielding quantitative scores with accompa-nying qualitative explanations, and improving the reliability and depth of the assessments.

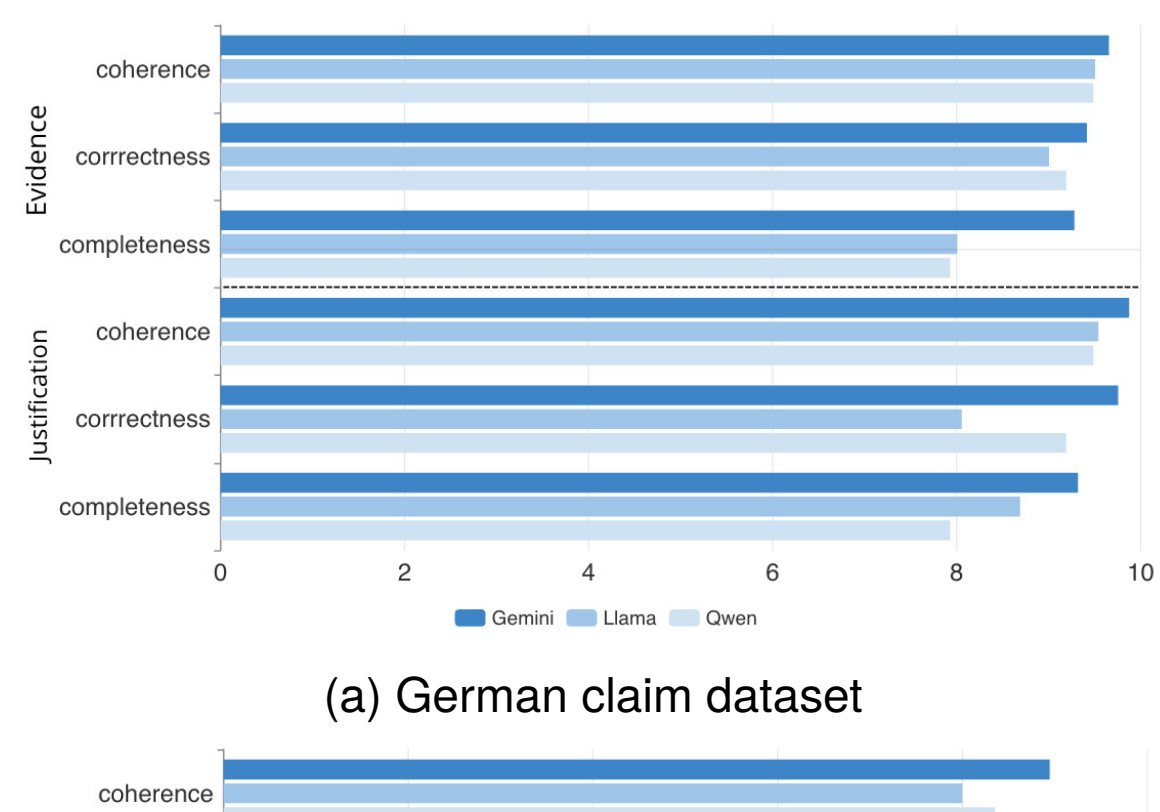

(a) German claim dataset

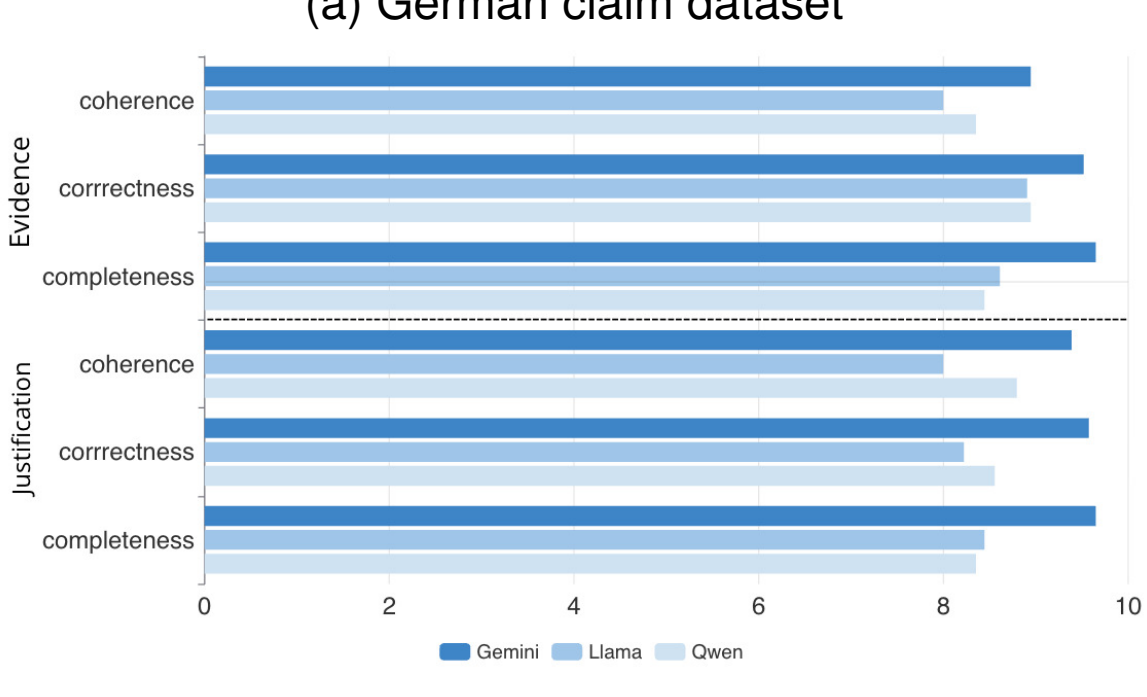

(b) French claim dataset

Figure 7: Evaluation scores of human annotators for evidence extraction and justification generation.

### 6.3. Evaluating Evidence Extraction and Justification Generation

We evaluated the performance of the structured text-based evidence extraction and the justification generation tasks across both languages and LLM configurations, using 50 randomly sampled claim–article pairs (average around 750 tokens per sample) for each language. For evidence extraction, Table 2 shows that across both datasets, the Gemini2.5-pro model achieves the highest G-Eval scores on all evaluation criteria. The relative ranking of Qwen2.5 and Llama3.3 varies across evaluation dimensions: LLama3.3 ranks second in coherence, whereas Qwen2.5 attains higher scores in correctness and completeness. These results indicate that Gemini2.5-pro retrieves and structures evidence information from debunking articles more consistently across languages, while LLama3.3 and Qwen2.5 trade places depending on whether structural flow (coherence) or factual sufficiency (correctness/completeness) is emphasized. In justification generation, similarly Gemini2.5-pro achieved the highest performance across all metrics. Qwen2.5 consistently reached the second highest scores, with Llama3.3 ranking third. This observed hierarchy suggests that Gemini2.5-pro generates justifications that are more contextually aligned and information-complete, with Qwen2.5 demonstrating proximate capabilities and Llama3.3 exhibiting a modest performance lag across all established criteria.

**Preliminary work: Human evaluation.** Three human annotators, two for German and one for French, evaluated the same subset per language used for the LLM evaluation for both tasks: evidence retrieval and justification generation. Following a detailed annotation guideline (see Appendix Fig. 8) and a fine-grained, penalty-aware rubric, they assessed each output for correctness, coherence, and completeness. Across both languages, the annotations revealed a consistent hierarchy of models, with Gemini achieving the strongest performance on both tasks (see Fig. 7). These findings align with the results obtained using the LLM-as-a-judge evaluation framework (see Table 2), supporting its reliability for model comparison.

### 6.4. Evaluating Multimodal LLMs: Visual Evidence Integration

In the second phase, we examined how incorporating visual evidence (images, their captions, and video keyframes with timestamps) affects the quality of justification generation. To isolate this effect, we created a text-only condition by removing all evidence items labeled under the multimedia evidence category from the generated text-based evidence dataset, and compared justifications produced in this condition to those generated when visual data were included. For consistency, we generated justifications using text evidence derived from the Gemini model's text-based evidence outputs, the top-performing system overall, and compared results across 200 French and 200 German claim samples. Under the text-only configuration, both models demonstrate strong and comparable performance across all evaluation dimensions for French and German claimReview data (see Table 3). In the French subset, Qwen 2.5-VL slightly surpasses Gemini2.5-pro in coherence and completeness, while Gemini reaches a marginally higher score in correctness. For the German subset, Gemini2.5-pro yields higher coherence, whereas Qwen achieves higher correctness and completeness. When evaluated on datasets incorporating multimodal evidence, Gemini2.5-pro consistently outperforms Qwen2.5-VL across all three criteria and both languages. These results indicate that Gemini generates justifications with stronger multimodal integration, yielding higher coherence, correctness, and completeness when reasoning over multimodal evidence, while Qwen shows weaker alignment between textual and visual signals, consistent with lower performance on these criteria.

## 7. Discussion and Conclusion

We proposed a unified, multilingual, multimodal pipeline which we used to produce datasets for French and German fact-checked claims, claim relevant metadata, normalized claim verdicts, aligned visual content, LLM-extracted standardized evidence categories and generated justifications.

We would like to stress that our goal was to use LLMs not as fact-checkers, but (i) as evidence-extraction tools based on structured evidence representations, and (ii) as generators of justifications from those evidences. Because these sources are grounded in the complex evidence structures of journalistic practices, taking them into account enables a more transparent LLM reasoning process. We believe our approach and our datasets will pave the way for future research on multilingual, multimodal fact-checking systems and cross-cultural and cross-organizational analyses of verification practices, revealing how strategies vary across media systems and contexts.

## 8. Limitations

Reliance on LLMs' judgments may introduce misinterpretations or inaccuracies in the evaluation of results, it is thus crucial to evaluate the relevance of their judgment. This is why we conducted an evaluation based on human validation of a substantial sample of LLM outputs. Our results demonstrate a strong correlation between our human annotators and LLM assessments.

A limitation of most fact-checking data collection is their dependence on the judgments of established fact-checking organizations. While these organizations are widely recognized, their assessments may reflect underlying biases related to their political orientation or their socio-cultural context. A comparative analysis involving multiple fact-checking organizations with competitive perspectives would be interesting. We leave it to future work.

## 9. Ethics statement

To evaluate the LLMs' ability to extract fact-checking evidence in a multimodal context, we collected data for which usage rights were not explicitly granted. As this research is publicly funded and not intended for commercial purposes, data collection is covered under the GDPR research exemption (Article 89)[3]. Furthermore, we do not release the videos or article content themselves, only links to the original sources.

Human annotators were paid according to standard rates as (student) research assistants for their contributions. We also made sure that the annotators were not exposed to any content in the articles that would be potentially offensive or harmful to them.

This work has been approved by our institutional review board (IRB).

## Acknowledgements

This work was partially supported by Djamé Seddah's chair at the PRAIRIE Institute, funded by the French National Research Agency. This project also received funding through the BPI Code Common and Scribe projects.

The authors gratefully acknowledge the computing time made available to them on the high-performance computer Noctua 2 at the NHR Center Paderborn Center for Parallel Computing (PC2). This center is jointly supported by the Federal Ministry of Research, Technology and Space and the state governments participating in the National High-Performance Computing (NHR) joint funding program[4].

## Bibliographical References

[3] https://gdpr-info.eu/art-89-gdpr/

[4] www.nhr-verein.de/en/our-partners

# A. Appendix

## A.1. Visual Media Collection from Debunking Articles

We collected visual media directly from the debunking articles associated with each claim statement for multimodal analysis. For each article, we scraped images, captions and video links using publisher-specific extraction pipelines. Table 4 summarizes the visual media collected across the French and German claim datasets. In total, the French dataset contains 10,557 images, and 1,361 video links spanning 3,357 and 1,047 articles respectively, while the German dataset contains 5,938 images and 194 video links.

| claimReview Datasets | Articles w/ Images | Total Images | Articles w/ Videos | Total Video Links |
|---|---|---|---|---|
| French | 3357 | 10557 | 1047 | 1361 |
| German | 2031 | 5938 | 172 | 194 |

Table 4: Visual media collected from debunking articles across the French and German claimReview datasets

## A.2. Evidence Category Analysis

To examine the association between categorical variables and evidence categories, we performed per-category Pearson's $\chi^2$ tests across publishers and labels. As shown in Appendix Tables 6, and 7, all per-category tests yielded statistically significant associations at the publisher and the label level. Standardized residuals (SR) were inspected to identify specific publisher-evidence and label-evidence combinations driving the overall significance. SR values exceeding 1.96 were interpreted as indicating statistically significant deviations from expected frequencies ( $p < 0.05$). Table 6 further reveals cross-dataset deviations at the publisher level. For example, *lemonde.fr* exhibits significantly low expert testimony (SR = -5.13) and multimedia usage (SR = -7.26), while *dpa-factchecking.com* (DE) shows significantly low quantitative evidence (SR = -7.08) and eyewitness accounts (SR = -5.40). At the label level, Table 7 shows that *FALSE*-labeled claims diverge across languages: in the French data they show elevated multimedia (SR = +6.19) and eyewitness evidence (SR = +5.59), whereas in the German data they exhibit significantly lower quantitative evidence (SR = -7.10) and eyewitness accounts (SR = -4.32).

| Datasets | Claim Count | Expert Testimony | | Quantitative Data & Statistics | | Official Records & Legal/Policy Documents | | Media Records (Mass or Social) | | Multimedia Evidence | | Eyewitness & Firsthand Accounts | |
|---|---|---|---|---|---|---|---|---|---|---|---|---|---|
| | | % | SR | % | SR | % | SR | % | SR | % | SR | % | SR |
| French claimReview data | 5170 | 75.1 | -1.40 | 81.4 | **+4.46** | 75.1 | $-0.57$ | 98.3 | $-0.11$ | 78.3 | **+3.41** | 36.7 | **+4.71** |
| German claimReview data | 3555 | 79.3 | +1.69 | 68.1 | **-5.38** | 76.8 | +0.69 | 98.7 | +0.13 | 68.2 | **-4.12** | 27.4 | **-5.68** |

Table 5: Relative percentages (%) and SR values for evidence category usage across language datasets. Per-category $\chi^2$ tests: Expert $\chi^2(1) = 20.53$, Quantitative $\chi^2(1) = 202.50$, Official $\chi^2(1) = 3.21$ ($p = .073$), Media $\chi^2(1) = 1.58$ ($p = .209$), Multimedia $\chi^2(1) = 110.29$, Eyewitness $\chi^2(1) = 80.84$; all $p < 0.001$. SR values in bold indicate $|SR| > 1.96$ ($p < 0.05$).

| | Publishers | Claim Count | Expert Testimony | | Quantitative Data & Statistics | | Official Records & Legal/Policy Documents | | Media Records (Mass or Social) | | Multimedia Evidence | | Eyewitness & Firsthand Accounts | |
|---|---|---|---|---|---|---|---|---|---|---|---|---|---|---|
| | | | % | SR | % | SR | % | SR | % | SR | % | SR | % | SR |
| French claimReview data | *20minutes.fr* | 590 | 73.4 | -0.95 | 72.4 | -1.00 | 64.6 | **-3.12** | 99.7 | +0.29 | 82.5 | **+2.36** | 39.5 | **+2.79** |
| | *defacto-observatoire.fr* | 124 | 88.7 | +1.51 | 84.7 | +1.11 | 83.1 | +0.93 | 99.2 | +0.08 | 76.6 | +0.31 | 38.7 | +1.13 |
| | *dpa-factchecking.com* | 298 | 67.8 | -1.78 | 65.1 | **-2.15** | 66.8 | -1.78 | 100.0 | +0.27 | 74.2 | -0.01 | 25.8 | **-2.12** |
| | *factuel.afp.com* | 2134 | 80.1 | +1.72 | 86.8 | **+5.73** | 77.6 | +0.95 | 99.1 | +0.28 | 98.1 | **+12.81** | 41.2 | **+6.68** |
| | *fasocheck.org* | 75 | 78.7 | +0.18 | 80.0 | +0.40 | 84.0 | +0.82 | 97.3 | -0.10 | 76.0 | +0.18 | 25.3 | -1.14 |
| | *francetvinfo.fr* | 309 | 78.0 | +0.24 | 88.0 | **+2.43** | 86.1 | **+2.09** | 94.2 | -0.76 | 49.5 | **-5.04** | 23.9 | **-2.74** |
| | *lemonde.fr* | 448 | 55.6 | **-5.13** | 64.5 | **-2.78** | 61.4 | **-3.50** | 98.2 | -0.05 | 44.6 | **-7.26** | 30.6 | -0.85 |
| | *liberation.fr* | 66 | 75.8 | -0.10 | 78.8 | +0.26 | 71.2 | -0.42 | 100.0 | +0.13 | 75.8 | +0.15 | 42.4 | +1.35 |
| | *numerama.com* | 51 | 78.4 | +0.13 | 88.2 | +1.01 | 66.7 | -0.75 | 96.1 | -0.17 | 92.2 | +1.49 | 23.5 | -1.17 |
| | *observalgerie.com* | 13 | 46.2 | -1.26 | 46.2 | -1.23 | 69.2 | -0.27 | 100.0 | +0.06 | 30.8 | -1.82 | 23.1 | -0.62 |
| | *science.feedback.org* | 15 | 100.0 | +1.02 | 93.3 | +0.77 | 93.3 | +0.78 | 100.0 | +0.06 | 86.7 | +0.56 | 13.3 | -1.32 |
| | *tf1info.fr* | 1047 | 73.4 | -1.24 | 85.1 | **+3.39** | 79.8 | +1.48 | 96.8 | -0.56 | 59.9 | **-5.37** | 36.6 | **+2.08** |
| German claimReview data | *apa.at* | 125 | 84.8 | +1.02 | 68.8 | -0.92 | 78.4 | +0.34 | 96.0 | -0.28 | 47.2 | **-3.50** | 17.6 | **-2.98** |
| | *br.de* | 141 | 90.1 | +1.80 | 74.5 | -0.20 | 81.6 | +0.79 | 84.4 | -1.68 | 31.2 | **-5.93** | 29.1 | -0.79 |
| | *correctiv.org* | 1398 | 80.9 | +1.74 | 75.7 | -0.12 | 76.0 | +0.12 | 100.0 | +0.58 | 70.7 | -1.53 | 27.5 | **-3.49** |
| | *dpa-factchecking.com* | 975 | 75.8 | -0.36 | 56.2 | **-7.08** | 78.2 | +0.86 | 99.5 | +0.32 | 63.1 | **-4.03** | 23.0 | **-5.40** |
| | *faktencheck.afp.com* | 489 | 83.8 | +1.77 | 79.6 | +0.91 | 79.3 | +0.91 | 100.0 | +0.34 | 99.6 | **+6.52** | 38.2 | **+2.06** |
| | *presseportal.de* | 241 | 72.2 | -0.82 | 49.8 | **-4.66** | 80.1 | +0.77 | 100.0 | +0.24 | 53.1 | **-3.80** | 23.7 | **-2.50** |
| | *rtl.de* | 19 | 57.9 | -0.94 | 42.1 | -1.69 | 42.1 | -1.69 | 100.0 | +0.07 | 68.4 | -0.29 | 57.9 | +1.90 |
| | *stern.de* | 47 | 70.2 | -0.52 | 42.6 | **-2.63** | 29.8 | **-3.62** | 83.0 | -1.07 | 83.0 | +0.70 | 27.7 | -0.63 |
| | *tagesschau.de* | 67 | 88.1 | +1.05 | 71.6 | -0.41 | 80.6 | +0.45 | 100.0 | +0.13 | 40.3 | **-3.22** | 29.9 | -0.43 |
| | *volksverpetzer.de* | 31 | 71.0 | -0.37 | 61.3 | -0.94 | 58.1 | -1.13 | 96.8 | -0.09 | 80.6 | +0.42 | 38.7 | +0.56 |
| | *wdr.de* | 22 | 31.8 | **-2.41** | 90.9 | +0.80 | 72.7 | -0.16 | 72.7 | -1.22 | 4.5 | **-3.79** | 13.6 | -1.57 |

Table 6: Relative percentages (%) and SR values for publishers-evidence categories across German and French datasets. Per-category $\chi^2$ tests: Expert $\chi^2(22) = 256.01$, Quantitative $\chi^2(22) = 637.90$, Official $\chi^2(22) = 230.78$, Media $\chi^2(22) = 480.10$, Multimedia $\chi^2(22) = 1670.08$, Eyewitness $\chi^2(22) = 217.26$. SR values in bold indicate $|SR| > 1.96$ ($p < 0.05$).

| Claim Labels | | Claim Count | Expert Testimony | | Quantitative Data & Statistics | | Official Records & Legal/Policy Documents | | Media Records (Mass or Social) | | Multimedia Evidence | | Eyewitness & Firsthand Accounts | |
|---|---|---|---|---|---|---|---|---|---|---|---|---|---|---|
| | | | % | SR | % | SR | % | SR | % | SR | % | SR | % | SR |
| French claimReview data | partially-true | 199 | 85.4 | +1.39 | 85.9 | +1.61 | 88.4 | **+2.06** | 98.0 | -0.07 | 59.8 | **-2.36** | 30.2 | -0.67 |
| | FALSE | 4482 | 74.6 | -1.70 | 80.3 | **+3.29** | 73.4 | -1.83 | 98.9 | +0.30 | 82.2 | **+6.19** | 37.7 | **+5.59** |
| | TRUE | 476 | 76.1 | -0.19 | 90.1 | **+3.54** | 85.7 | **+2.50** | 93.1 | -1.19 | 49.4 | **-6.29** | 29.4 | -1.32 |
| | other | 13 | 61.5 | -0.63 | 76.9 | +0.04 | 61.5 | -0.59 | 92.3 | -0.22 | 84.6 | +0.44 | 46.2 | +0.83 |
| German claimReview data | partially-true | 696 | 86.1 | **+2.78** | 84.9 | **+2.71** | 86.8 | **+3.34** | 99.1 | +0.18 | 60.8 | **-4.11** | 23.7 | **-4.23** |
| | FALSE | 2699 | 78.7 | +1.12 | 64.1 | **-7.10** | 75.1 | -0.39 | 98.9 | +0.22 | 70.4 | **-2.31** | 28.1 | **-4.32** |
| | TRUE | 102 | 66.7 | -1.17 | 76.5 | +0.06 | 71.6 | -0.49 | 89.2 | -0.94 | 60.8 | -1.57 | 33.3 | +0.08 |
| | other | 58 | 48.3 | **-2.48** | 39.7 | **-3.17** | 43.1 | **-2.86** | 100.0 | +0.12 | 72.4 | -0.16 | 29.3 | -0.48 |

Table 7: Relative percentages (%) and SR values for claim labels–evidence categories across German and French datasets. Per-category $\chi^2$ tests: Expert $\chi^2(7) = 93.84$, Quantitative $\chi^2(7) = 390.19$, Official $\chi^2(7) = 139.67$, Media $\chi^2(7) = 164.94$, Multimedia $\chi^2(7) = 419.48$, Eyewitness $\chi^2(7) = 105.73$; all $p < 0.001$. SR values in bold indicate $|SR| > 1.96$ ($p < 0.05$).

## A.3. Inter-annotator Agreement Scores

For the German claims, we computed inter-annotator agreement between two annotators across six evaluation dimensions: evidence correctness, evidence coherence, evidence completeness, justification correctness, justification coherence, and justification completeness. Each dimension was assessed for three model outputs (Gemini, Qwen, and Llama), and the reported agreement scores were computed on aggregate dimension scores obtained by averaging the three model-specific ratings for each annotator. Ratings were assigned on a 0-10 scale in increments of 0.5, with 10 indicating the highest quality.

Table 8 reports Krippendorff's $\alpha$ (interval), Gwet's AC2 (quadratic weights), and the Absolute Annotation Difference (AAD). Krippendorff's $\alpha$ ranged from -0.043 (justification coherence) to 0.419 (justification correctness), with an overall mean of 0.157. Gwet's AC2 was consistently higher, ranging from 0.653 to 0.877 (overall: 0.764). This divergence is consistent with the known sensitivity of $\alpha$ to restricted score ranges and low between-item variance, whereas AC2 often yields more stable values under skewed or concentrated rating distributions.

To complement these chance-corrected coefficients, we report AAD, which measures the per-instance absolute difference between annotators' aggregate scores. The overall mean AAD was 0.84 on the aggregated 0-10 score scale (median: 0.67), with 73% of annotation pairs differing by at most 1.0 point and 93% by at most 2.0 points. Even for justification coherence ($\alpha = -0.043$), the mean AAD was only 0.46, with 92% of pairs within 1.0 point. Taken together, these results suggest that the low $\alpha$ values are more consistent with restricted score variance and small practical score differences than with large substantive disagreement between annotators.

## A.4. Annotation Guideline and Setup

To prevent annotators from applying different standards based on their prior expectations of a given LLM, we anonymized all model outputs. We assigned the three models (Gemini, Qwen, and Llama) neutral labels (Model-A, Model-B, Model-C) and varied the label-to-model assignment across instances, so that annotators could not infer model identities through repeated exposure. The mapping between labels and model identities was revealed only after all annotation was completed. Before the main annotation phase, we conducted a calibration workshop in which we walked the annotators through the guideline (Fig. 8), the scoring rubric, and we worked through examples from the claim subset together. With this procedure, our objective was to establish a shared interpretive baseline and to reduce systematic divergence in how the rubric was applied across tasks and models.

| Annotation Category | | Krippendorff's $\alpha$ | Gwet's AC2 | AAD Mean | AAD Median | AAD $\% \leq 1.0$ | AAD $\% \leq 2.0$ |
|---|---|---|---|---|---|---|---|
| Evidence | correctness | 0.073 | 0.734 | 0.75 | 0.67 | 76 | 94 |
| | coherence | 0.248 | 0.877 | 0.48 | 0.33 | 90 | 98 |
| | completeness | 0.138 | 0.653 | 1.30 | 1.17 | 46 | 84 |
| Justification | correctness | 0.419 | 0.764 | 0.73 | 0.67 | 76 | 94 |
| | coherence | -0.043 | 0.822 | 0.46 | 0.33 | 92 | 100 |
| | completeness | 0.108 | 0.733 | 1.32 | 1.00 | 56 | 88 |
| OVERALL | | **0.157** | **0.764** | **0.84** | **0.67** | **73** | **93** |

Table 8: Inter-annotator agreement scores for German claim evaluation across six dimensions

We are trying to evaluate whether large language models (LLMs) can be helpful in fact-checking claims. For this, we have data in the form of factual claims which have been checked by news organizations who have then published a “debunking article” discussing evidence for and against the truth of the claim.

This annotation study assesses and compares the outputs of three large language models on two tasks: (i) **evidence extraction** from debunking articles and (ii) **justification generation** for given “claim / evidence ” pairs. The study is conducted in a realistic fact-checking setting using real-world claims, the ratings issued by professional fact-checking organizations, and the corresponding debunking articles, which serve as the sole authoritative source.

You will evaluate each model’s evidence and justification independently based on the claim, its assigned label and relevant article using three criteria: coherence (clear structure, logical flow, and consistent formatting), correctness (faithful to the article, correct categorization, no hallucinations, and label-consistent reasoning), and completeness (inclusion of all salient evidence types and key items, and full coverage of the reasoning steps needed to support the label). The aim is to produce reliable, comparable quality judgments across the three models.

Each case provides:

(a) a **Claim**, i.e., a factual statement that may be true or false;

(b) its **Debunking Article (URLs)**, which serves as the sole source of truth;

(c) a given **Claim Label** (True / Partially true / False / Other);

**True –** The claim is fully accurate as stated; no essential context is missing, and nothing in the presentation is misleading.

**Partially true –** The claim falls on a spectrum of truth, containing elements that are both true and false.

**False –** The claim is clearly incorrect or is directly contradicted and debunked by the debunking article.

**Other –** The claim is opinion-based, satirical, and unverifiable.

(d) the extracted *Evidence* by three models based on the Claim and Debunking Article, and their predefined evidence categories;

**Expert Testimony** — refers to statements or analyses whose credibility comes from the recognized knowledge, training, or authority of a professional in a relevant field. This includes interviews, commentary, and judgments made by professors, doctors, scientists, engineers, lawyers, and other licensed specialists, as well as official spokespersons acting explicitly in an expert capacity (such as a CDC epidemiologist explaining disease spread). The guiding principle is that if the evidential value lies in the expert’s interpretation, analysis, or professional judgment, it should be classified as Expert Testimony.

**Quantitative Data & Statistics** — covers evidence in the form of measurable figures used as proof. This includes percentages, rates, margins of error, poll results, survey data, financial statements, budgets, and model estimates such as forecasts or projections. The decision rule is when the evidential value lies in the numbers themselves, rather than in someone’s interpretation, it should be classified as Quantitative Data & Statistics.

**Official Records & Legal/Policy Documents** — are formal and authoritative materials issued by governments, institutions, or organizations in an official capacity. This category encompasses laws, regulations, treaties, contracts, court filings, judicial rulings, governmental and institutional reports, etc.

**Media Record (Mass or Social)** — covers publications from journalism and non-governmental social media accounts that serve as evidence of the publication or circulation of a claim or piece of information through mass or social media, such as news articles, editorials, blogs, newsletters, online forums, and social media posts.

**Multimedia Evidence** — Multimedia Evidence includes all forms of visual, audio, or interactive files used as proof. This category covers photos, videos, audio recordings, screenshots of posts, comments, charts, or documents, as well as embedded social media posts when the media object itself is the cited evidence. Maps, diagrams, and graphs presented as images also qualify.

**Eyewitness & Firsthand Accounts** — captures evidence whose credibility derives from direct personal observation or lived experience of the event in question, as distinct from institutionally recognized expertise.

(e) the generated *Justification* by three models based on the claim, extracted evidence, and logically supports the given claim label.

Each criterion (coherence, completeness, correctness) should be scored on a 0–10 scale with a step size of 0.5:

- **0 = very poor** (fails the requirement)
- **5 = partial** (some correct elements, but noticeable problems)
- **10 = excellent** (fully meets the requirement)

The outputs of *Model-A, Model-B,* and *Model-C* on the two tasks; evidence extraction and justification generation; must be evaluated using this continuous scale. Evaluators are expected to make full use of the entire 0–10 range rather than defaulting to the midpoint. Scores of 0, 5, and 10 serve as anchors, while intermediate values (e.g., 3, 7, 9) should be applied to capture in-between quality. Consistent application of the criteria across tasks and models is required to produce reliable and comparable results.

### 1. How to Evaluate the Evidence Extraction

Please evaluate how well the model has extracted the evidence for or against the claim from the debunking article.

The evidence must be organized under the six predefined types. If the article does not contain a given type of evidence, leaving that category empty is correct. Items should follow a consistent format and be easy to scan. Locators such as figure identifiers or timestamps are strongly preferred as they improve traceability.
Penalize outputs that mix categories, duplicate the same fact across categories without a clear justification (Cross-referencing is allowed and expected), or employ an inconsistent structure that impairs readability.

**0** = Categories are mixed, inconsistent, hard to read; duplications or messy format
**5** = Categories mostly respected, but some mixing, duplication, or structural inconsistencies reduce readability
**10** = Perfectly organized under the six predefined categories, consistent format, easy to scan, with locators (timestamps, person affiliation) whenever possible

#### 1a. Correctness – Evidence (Fidelity and Categorization)

Each evidence item must appear in the article as either a short quote or a faithful paraphrase, preserving details such as dates, locations, and qualifiers. Items must be placed under the correct evidence type according to the definitions above. There must be no hallucinations, such as invented names, numbers, media, or documents.
Penalize misquotations, invented or altered statistics, incorrect category assignments, and paraphrases that change the meaning or scope of the original content.

**0** = Contains hallucinations, misquotations, or wrong categories
**5** = Mostly faithful, but some items are misplaced or inaccurately paraphrased
**10** = All evidence is faithfully quoted/paraphrased, correctly categorized, with details (dates, locations, qualifiers) preserved

#### 1b. Completeness – Evidence (Coverage)

Whenever the article includes a particular type of evidence, the corresponding category must contain the relevant items. The extraction should include the key pieces of evidence (for example, central quotations, pivotal statistics, core documents, critical media, and major eyewitness accounts). Evidence should not be misplaced in a way that obscures its relevance.
**Note:** Leaving a category blank is correct if the article contains no evidence of that type.

**0** = Misses most key evidence
**5** = Includes some central evidence, but omits important items
**10** = Fully covers all key evidence when present; categories left blank only when truly absent

### 2. How to Evaluate the Justification Generation

Please evaluate the quality of the provided justification for the claim label (true/partially true/false/other) based on the claim statement and extracted evidence.

#### 2a. Coherence – Justification (French/German Flow and Logic)

The justification must be written in fluent, natural French and should form a coherent paragraph (or a tightly connected set of sentences). It should use clear logical connectors and present a transparent progression from the claim to the cited evidence and finally to the label conclusion. References to specific evidence items must be clear and consistent.

**0** = Incoherent, ungrammatical, or illogical
**5** = Understandable but rough; weak connectors or fragmented reasoning
**10** = Fluent French, clear logical flow: claim $\rightarrow$ evidence $\rightarrow$ conclusion

#### 2b. Correctness – Justification (Alignment with Evidence and Label)

The justification must rely exclusively on the extracted evidence and must not introduce external facts or speculation. It should interpret the evidence accurately and avoid contradictions. The reasoning must support the given label according to the definitions provided.
Label cues:

*True:* The justification demonstrates that the claim aligns with the article's evidence and that no critical context is missing.

*Partially true:* The justification explicitly distinguishes between what is accurate and what is misleading, incomplete, or incorrect.
*False:* The justification clearly shows that the claim is contradicted or debunked by the evidence.
*Other:* The justification explains that the claim is opinion-based, satire, and unverifiable.

**0** = Contradicts evidence, adds speculation, or misinterprets facts
**5** = Generally aligned but with weak or partial interpretation
**10** = Fully consistent with extracted evidence and the assigned label

### 2c. Completeness – Justification (Reasoning Coverage)

The justification must present the key reasoning steps that connect the claim to the evidence and from the evidence to the assigned label. It should address salient counter-evidence when the article provides it. The argumentation should not skip logical steps and should avoid bare assertions that lack evidentiary support.

**0** = Bare assertions, skips key reasoning, ignores counter-evidence
**5** = Some reasoning, but missing steps or counterpoints
**10** = Covers all reasoning steps, addresses counter-evidence when present, connects claim → evidence → label without gaps

## 3. Step-by-Step Annotation Workflow

1. Read the claim closely and examine the linked article to determine the key facts, catalogue the evidence types present, and map their connection to the claim.
2. Check the model's Evidence Section:
   - Verify each item exists in the article.
   - Verify correct categorization (put in the right evidence type).
   - Check for missing major items or categories that the article clearly contains.
3. Evaluate Evidence Dimensions:
   - Coherence – Evidence: Is the evidence section well-organized, consistent, and readable across categories?
   - Correctness – Evidence: Is every evidence item faithful to the article and placed in the right category?
   - Completeness – Evidence: Did the model include all major evidence and all relevant categories present in the article?
4. Read the Justification:
   - Must be natural, fluent French.
   - Must use only the extracted evidence (no outside facts).
   - Must clearly tie the evidence to the given label.
5. Evaluate Justification Dimensions:
   - Coherence – Justification: Is the French explanation well-structured, flows logically, and easy to follow?
   - Completeness – Justification: Does it cover all key reasoning steps and address the major points needed to assess the label?
   - Correctness – Justification: Is it consistent with the evidence and the label, with no hallucinations or contradictions?

## 4. Deduction Rules

### Deduction rules for extracted evidence

| Task | Sub-criterion | Error / Condition | Deduction (per occurrence) | Max deduction |
|---|---|---|---|---|
| Evidence | Coherence | Mixing category names/structures | $-2$ | $-4$ |
| Evidence | Coherence | Duplicating the same fact across categories | $-1$ | $-3$ |
| Evidence | Coherence | Readable, clear items | $-1$ | $-3$ |
| Evidence | Correctness | Paraphrase/quote meaning drift | $-0.5$ | $-2$ |
| Evidence | Correctness | Mis-categorization of an item or including the justification part as evidence | $-1$ | $-3$ |
| Evidence | Correctness | Broken/mismatched cross-reference | $-0.5$ | $-1.5$ |
| Evidence | Correctness | Misattributed quote/media/document, unsupported/altered number | $-2$ | $-4$ |
| Evidence | Correctness | Hallucinated entity/quote/media/document | $-2.5$ | $-5$ |
| Evidence | Completeness | Omitted a major evidence item present in the article | $-1.5$ | $-4.5$ |
| Evidence | Completeness | Missing locators for major items (timestamp, figure ID, affiliation) | $-1$ | $-4$ |
| Evidence | Completeness | Partial capture of a multi-part argument | $-0.5$ | $-2$ |

### Deduction rules for generated justification

| Task | Sub-criterion | Error / Condition | Deduction (per occurrence) | Max deduction |
|---|---|---|---|---|
| Justification | Coherence | Missing/weak logical connectors | $-0.5$ | $-3$ |
| Justification | Coherence | Disjointed reference to evidence (unclear mapping) | $-1$ | $-4$ |
| Justification | Coherence | Awkward/unidiomatic French/German hindering readability | $-0.5$ | $-3$ |
| Justification | Correctness | Uses facts not in extracted evidence | $-2$ | $-4$ |
| Justification | Correctness | Misinterprets arguments/qualifier/date/number from evidence and overreaching | $-2$ | $-4$ |
| Justification | Correctness | Conclusion inconsistent with label definition | $-2$ | $-2$ |
| Justification | Completeness | Skips key step (claim→evidence or evidence→label) | $-1$ | $-3$ |
| Justification | Completeness | Ignores salient counter-evidence present | $-2$ | $-4$ |
| Justification | Completeness | Bare assertion without citing a supporting extracted item | $-1$ | $-3$ |

## 5. Final Checklist

### Evidence extraction:

- **Coherence score (0–10):**
  - Comment on organization into six categories: Are empty categories left blank?
  - Note if bullets are concise and consistently formatted.
  - Check whether locators (dates, sources, identifiers) are present.
  - Identify any issues: category mixing, inconsistent formatting.
- **Correctness score (0–10):**
  - State whether all items are directly present in the article, with scope preserved (dates, places, qualifiers).
  - Check for hallucinations or invented details.
  - Assess whether categorization matches the definitions (e.g., stats under Quantitative Data & Statistics, expert statements under Expert Testimony).
  - Identify any misquotes or paraphrase errors.
- **Completeness score (0–10):**
  - List the main strands of evidence in the article (e.g., expert quotes, statistics, provenance, documents).
  - Check whether each is captured under the correct category.
  - Comment on whether any pivotal items are missing or misplaced.
  - Note if blank categories are justified.

### Justification generation:

- **Coherence score (0–10):**
  - Comment on fluency of French/German: smooth/awkward?
  - Check logical flow: claim → evidence → conclusion.
  - Note use of connectors ("mais", "cependant", "en revanche", etc.).
  - Identify any disjointed or unclear sentences.
- **Correctness score (0–10):**
  - State whether only extracted evidence was used; check no external facts/speculation.
  - Verify accurate interpretation of evidence (dates, numbers, qualifiers preserved).
  - Confirm the conclusion aligns with the assigned label definition.
  - Flag any misrepresentation, contradiction, or hallucination.
- **Completeness score (0–10):**
  - Check if all major evidence strands cited in the extraction are represented in the reasoning.
  - Ensure clear link from evidence to conclusion — no skipped logic.
  - Note if salient counter-evidence was considered or ignored.
  - Assess whether the argument avoids bare assertion and is fully explained.

Figure 8: Annotation guideline for evaluating LLM performance on evidence extraction and justification generation.